%% file: main.tex
\documentclass[times,preprint,10pt]{elsarticle}
\input{preamble}
\begin{document}
\input{frontmatter}
\input{paper_content}
\end{document}

%% file: preamble.tex
\usepackage{amsmath,amssymb,graphicx,booktabs,longtable,array,multirow,microtype}
\usepackage{setspace,caption,placeins}
\usepackage[hidelinks]{hyperref}
\hypersetup{
  pdftitle={LAMP: Layerwise Adaptive Mixing of Predictors for Training-Free Few-Shot Vision-Language Adaptation},
  pdfauthor={Zhipeng Ye, Feng Jiang, Qiufeng Wang, Hao Li},
  pdfsubject={Few-shot vision-language adaptation},
  pdfkeywords={Few-shot classification, Vision-language models, Training-free adaptation, Kernel residual learning, Layerwise predictor mixing}
}
\biboptions{numbers,sort&compress}
\myfooter[C]{\thepage}

\newcommand{\Normalize}{\operatorname{norm}_2}

\ifdefined\pdfminorversion
\fi

%% file: frontmatter.tex
\begin{frontmatter}
\title{LAMP: Layerwise Adaptive Mixing of Predictors for Training-Free Few-Shot Vision--Language Adaptation}
\author[a]{Zhipeng Ye}
\ead{zhipengye@nustti.edu.cn}
\author[a]{Feng Jiang\corref{cor1}}
\ead{jf@nustti.edu.cn}
\author[b]{Qiufeng Wang}
\ead{qiufeng.wang@xjtlu.edu.cn}
\author[c]{Hao Li}
\ead{lihao@arizona.edu}
\cortext[cor1]{Corresponding author.}
\address[a]{Taizhou Institute of Science and Technology, Nanjing University of Science and Technology, Taizhou 225300, Jiangsu, China}
\address[b]{Department of Intelligence Science, Xi'an Jiaotong-Liverpool University, Suzhou 215123, Jiangsu, China}
\address[c]{School of Computer Science and Technology, University of Arizona, Tucson 85705, AZ, USA}
\begin{abstract}
Few-shot adaptation often relies on an encoder's final global representation, leaving complementary information in intermediate layers underused. Combining these features is challenging because their dimensions can differ and their usefulness varies across tasks. We propose LAMP, Layerwise Adaptive Mixing of Predictors, a training-free framework that combines global and regional information from multiple encoder layers. A single forward pass per view supplies all descriptors. Global and support-calibrated regional kernel predictors independently fit the same label residuals. Labeled validation data select their mixture in class-score space, preserving native feature dimensions without learned projections. LAMP-F further adapts text-anchored class prototypes, using a cached response to update the fixed-kernel residual predictions exactly. Experiments cover eleven datasets and ViT and ResNet backbones. LAMP achieves state-of-the-art mean accuracy among the compared published training-free ResNet-50 methods at 4, 8, and 16 shots, reaching 78.73\% at 16 shots. Averaged across five support budgets, LAMP improves over a paired global-feature-only baseline by 1.14 and 1.59 percentage points on ViT-B/16 and ResNet-50; LAMP-F raises these gains to 1.96 and 2.54 points. LAMP-F reaches 16-shot accuracies of 83.57\% and 79.10\%, respectively. Controlled experiments identify multi-layer integration as the main source of improvement and examine pooling, calibration, validation labels, and computational cost. These findings show that combining layer-specific predictions can exploit complementary features from frozen encoders to improve few-shot classification.
\end{abstract}
\end{frontmatter}
\thispagestyle{plain}

%% file: paper_content.tex
\input{introduction}
\input{related}
\begin{figure}[htbp]
\centering\includegraphics[width=\linewidth]{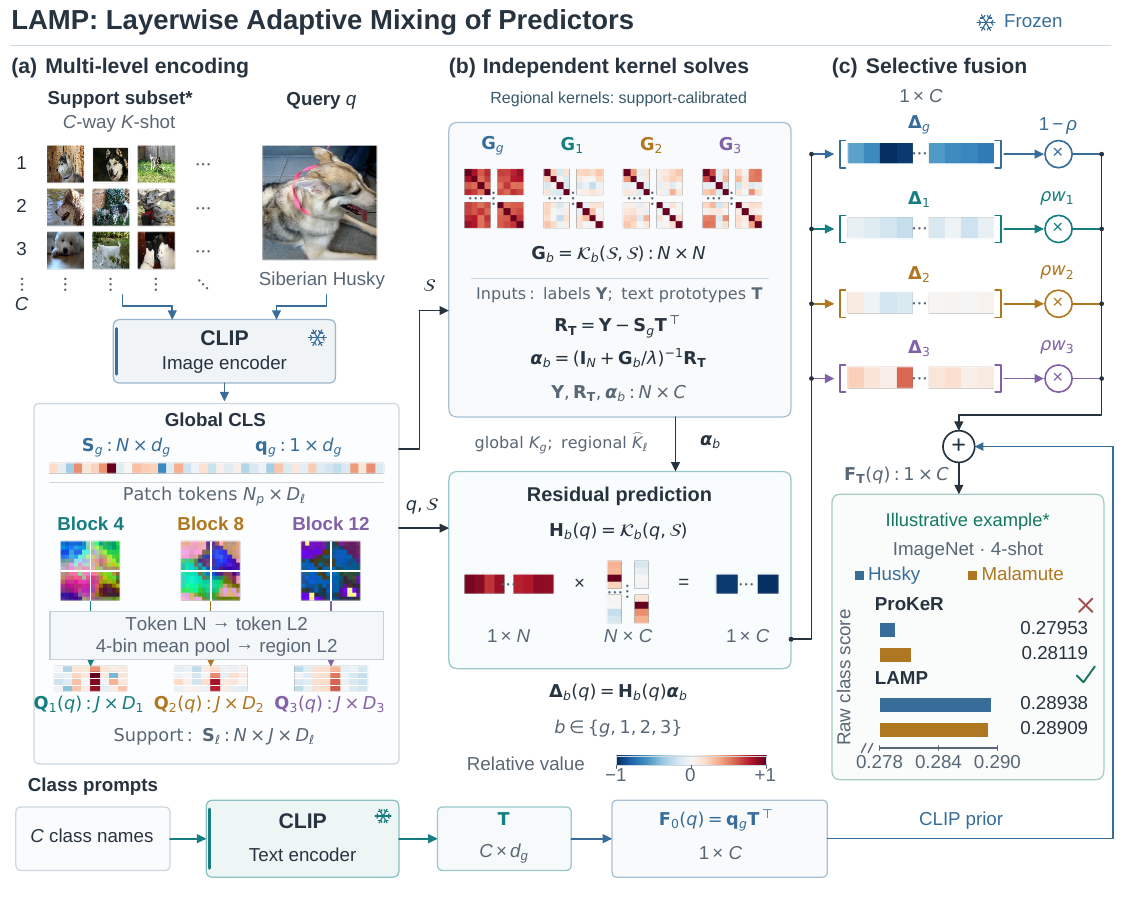}
\caption{Training-free LAMP with ViT-B/16 for one query $q$. (a) The frozen encoder provides global descriptors, query regional matrices $\mathbf{Q}_\ell(q)$, and support tensors $\mathbf{S}_\ell$; branches $\ell=1,2,3$ correspond to blocks 4, 8, and 12. Row labels $1,2,3,\ldots,C$ denote class indices. (b) Independent predictors fit the shared support-label residual $\mathbf{R}_{\mathbf{T}}$. The effective kernel $\mathcal K_b$ is $K_g$ for the global branch and $\widehat K_\ell$ for a calibrated regional branch. (c) Their corrections $\boldsymbol{\Delta}_b(q)$ are weighted and added to the CLIP prior. Here $N=VCK$, $V=10$, $J=4$, $D_\ell=768$, and $N_p=196$. Prediction uses the selected parameters; Section~\ref{sec:prototype-adaptation} describes LAMP-F. Asterisks mark the displayed support subset and illustrative prediction.}\label{fig:method}
\end{figure}
\input{method}

\input{experiments}
\FloatBarrier
\input{discussion}
\input{conclusion}
\input{backmatter}
\bibliographystyle{elsarticle-num}
\bibliography{references}

%% file: introduction.tex
\section{Introduction}

Vision--language models such as CLIP recognize images by matching visual representations to category descriptions~\cite{clip}. Few-shot adaptation uses a small labeled support set to improve this pretrained predictor for a target task. Many adaptation methods rely on the encoder's final global descriptor. This compact representation supports image--text matching, but intermediate token grids and convolutional maps also retain visual information at different levels of abstraction. These features are already available during the encoder's forward pass. Using them offers an opportunity to improve adaptation while keeping the encoder frozen.

Existing approaches use different sources of information to adapt the predictor. Prompt and prototype methods refine the class representations~\cite{coop,taskres,clap}. Tip-Adapter retrieves labeled instances, while ProKeR fits a regularized support-label residual; both use final global descriptors to match images~\cite{tipadapter,proker}. SSP instead uses local features to construct subspaces for image--text alignment~\cite{ssp}. These methods motivate the use of complementary semantic and visual information. Integrating features from several encoder layers, however, requires addressing two related challenges: their native dimensions can differ, and the most useful layers can vary with the target task.

Our central idea is to combine predictions from different layers in a shared class-score space. Different layers provide different measures of image similarity, but each predictor outputs a score for the same set of classes. Their predictions can therefore be combined while preserving the original feature dimensions. Fitting these predictors separately also allows their outputs to be reused when selecting the mixture, so each candidate mixture does not require a new fit.

Based on this idea, we introduce training-free \textbf{LAMP}, \textbf{L}ayerwise \textbf{A}daptive \textbf{M}ixing of \textbf{P}redictors. A single encoder pass per view yields a global descriptor and regional descriptors from three layers of a Vision Transformer (ViT) or ResNet~\cite{vit,resnet}. The global predictor and three regional predictors independently fit the same label residual: the difference between the support labels and CLIP's initial scores. Each regional kernel is calibrated against the support set, and labeled validation data select the mixture of global and regional corrections. LAMP thus uses complementary information from several layers without gradient optimization or learned feature projections. Its extension, LAMP-F, adapts text-anchored class prototypes while retaining the selected kernels and mixture weights. A cached linear response accounts exactly for the resulting changes to both the initial scores and residual corrections, avoiding repeated matrix factorization during prototype evaluation.

We evaluate LAMP on eleven datasets, five support budgets, and CLIP backbones from the ViT and ResNet families. In the comparison with published training-free ResNet-50 (RN50) methods, LAMP achieves state-of-the-art mean accuracy at 4, 8, and 16 shots, reaching 78.73\% at 16 shots. LAMP-F reaches 83.57\% and 79.10\% at 16 shots on ViT-B/16 and RN50, respectively. Across the eleven datasets and five support budgets, LAMP improves over our paired global-feature-only baseline by an average of 1.14 and 1.59 percentage points; LAMP-F raises these gains to 1.96 and 2.54 points. Controlled comparisons identify the integration of multiple encoder layers as the main source of improvement and examine the effects of regional pooling and calibration. Experiments with RN101 and ViT-B/32, together with transfer to ImageNet-V2 and ImageNet-Sketch, assess generalization across architectures and image distributions~\cite{imagenetv2,imagenetsketch}.

\begin{samepage}
Our contributions are:
\begin{itemize}
\renewcommand{\labelitemi}{$\bullet$}
\item We introduce LAMP, a training-free method that combines global and regional information from multiple encoder layers using independently fitted kernel predictors. Regional kernels are calibrated against the support set. Score-level fusion preserves native feature dimensions, and separating branch fitting from mixture selection makes predictions reusable across candidates.
\item We develop LAMP-F, which combines prototype adaptation with the same global and regional predictors. For fixed kernels and mixture weights, its cached response exactly updates both the initial class scores and residual corrections as the prototypes change.
\item We demonstrate state-of-the-art mean accuracy in the published training-free RN50 comparison at 4, 8, and 16 shots across eleven datasets. Paired comparisons on ViT and ResNet backbones establish the accuracy gains, and controlled experiments explain the contribution of multi-layer integration.
\end{itemize}
\end{samepage}

%% file: related.tex
\section{Related work}\label{sec:related}
\subsection{Class-level semantic adaptation}
Semantic adaptation adjusts a vision--language model's class representations to a target task. CoOp and CoCoOp learn contextual prompts, while KgCoOp and ProGrad constrain prompt updates to preserve pretrained knowledge~\cite{coop,cocoop,kgcoop,prograd}. MaPLe couples visual and textual prompts; PLOT and CAPEL use multiple prompts or class representations to capture variation within a category~\cite{maple,plot,chen2026capel}. TaskRes learns residual updates to class prototypes, while CLAP adapts prototypes with a text-anchoring penalty~\cite{taskres,clap}. These methods improve the class representations through optimization. LAMP keeps the class text vectors fixed and uses visual similarities to correct their prediction errors on the support set. Its trainable extension, LAMP-F, combines CLAP's text anchoring with global and regional residual predictors. Cached responses account for the effects of prototype updates on both the initial query scores and the support residuals, allowing prototype adaptation to work with the selected visual predictors.

\subsection{Instance-level global adaptation}
Instance-level adaptation draws on visual information from labeled support images. CLIP-Adapter learns transformations of global embeddings, and CLIP-LoRA updates internal attention weights through low-rank parameters~\cite{clipadapter,cliplora}. Both adapt visual representations through gradient optimization. Training-free methods instead adjust how frozen features are used for prediction: Tip-Adapter retrieves support labels, APE refines feature channels and affinities, and GDA estimates class statistics~\cite{tipadapter,ape,gda}. IDEA combines image similarities with similarities between image captions, and T-IDEA further learns alignment and bias correction~\cite{idea}. LAMP obtains complementary features from intermediate encoder layers and combines their predictions. This allows it to use information beyond the final global descriptor while preserving each layer's original feature space.

LAMP uses ProKeR's residual kernel formulation, which transfers corrections from support images to queries according to global-feature similarities~\cite{proker}. It fits a global predictor and separate regional predictors whose kernels are calibrated against the support set. Combining their outputs in class-score space preserves each branch's feature dimensions and similarity structure; labeled validation data determine their relative contributions. Since the branches are fitted independently, their predictions can be reused when evaluating different mixtures. LAMP-F also exploits their linear dependence on the support residual to cache the combined response to prototype updates.

\subsection{Patch- and region-level adaptation}
Selective Vision-Language Subspace Projection (SSP) uses local features to construct visual and class-specific language subspaces for image--text alignment without gradient optimization~\cite{ssp}. Both SSP and LAMP use information beyond a single global descriptor, but they integrate it differently. SSP improves cross-modal alignment through subspace projection. LAMP uses regional features from several encoder layers to fit the same support-label residual and then combines the resulting class scores. Because all branches predict the same classes, their outputs can be combined even when their feature dimensions differ. A single encoder pass per view provides all regional descriptors. LAMP uses generalized-mean pooling for convolutional activations~\cite{gem} and kernel-mean comparisons of descriptor sets~\cite{kme}. Comparisons with simpler summaries of the same layers distinguish the benefit of using multiple layers from that of retaining regional descriptor sets.

\FloatBarrier

%% file: method.tex
\section{Method}
\label{sec:method}

LAMP adapts CLIP by using labeled support images to correct its initial class scores. The support labels indicate which scores should increase or decrease, and visual similarity determines how these corrections transfer to a query. Global descriptors and regional features from intermediate layers capture different relationships between images, providing complementary ways to correct a prediction. LAMP combines the resulting corrections in class-score space: every branch produces one score per class, even when the branches have different feature dimensions.

We first introduce the residual target and global predictor using ProKeR's residual kernel formulation~\cite{proker}. We then explain how intermediate layers provide regional descriptors, how their similarities are calibrated, and how the resulting predictors fit the same target. Validation determines the contribution of each predictor. LAMP-F further adapts the class prototypes while keeping the selected visual kernels and mixture weights fixed. Figure~\ref{fig:method} illustrates the method with a ViT backbone.

\subsection{Residual prediction around the CLIP prior}
\label{sec:preliminaries}

Consider $C$ classes with $K$ labeled images each. For support image $i\in\{1,\ldots,CK\}$ with class label $y_i\in\{1,\ldots,C\}$, frozen CLIP~\cite{clip} encodes $V=10$ augmented views $x_{i,v}$, $v\in\{1,\ldots,V\}$. These views form the ordered support bank $\mathcal{S}=(s_1,\ldots,s_N)$, where $N=VCK$; each entry inherits its source image's label. The support budget therefore counts $CK$ labeled images, and the kernel bank contains $N$ augmented views. The ordered query collection $\mathcal{Q}=(q_1,\ldots,q_M)$ contains one deterministic view per query.

Vectors are column vectors unless stated otherwise; feature matrices contain their transposes as rows. In particular, $\mathbf{S}_g\in\mathbb{R}^{N\times d_g}$ and $\mathbf{Q}_g\in\mathbb{R}^{M\times d_g}$ contain unit global descriptors $\mathbf{g}(x)\in\mathbb{R}^{d_g}$. The matrix $\mathbf{Y}\in\{0,1\}^{N\times C}$ contains one-hot support-view labels, and $\mathbf{T}\in\mathbb{R}^{C\times d_g}$ contains unit class text vectors from fixed prompts as rows. For one query $q$, $\mathbf{q}_g=\mathbf{g}(q)^\top\in\mathbb{R}^{1\times d_g}$ is its row in $\mathbf{Q}_g$. We write $\Normalize$ for L2 normalization of a vector and row-wise L2 normalization of a matrix. Bold symbols denote vector-, matrix-, or tensor-valued quantities; calligraphic $\mathcal{S}$ and $\mathcal{Q}$ denote ordered collections.

The text matrix $\mathbf{T}$ defines both the initial query scores and the support scores used to form the residual target. Each CLIP score is the dot product of an image descriptor and a class text vector. For a support view, subtracting these scores from its one-hot label gives the desired correction for each class. All branches fit this same residual:
\begin{equation}
\mathbf{F}_0(\mathcal{Q})=\mathbf{Q}_g\mathbf{T}^\top,\qquad \mathbf{R}_{\mathbf{T}}=\mathbf{Y}-\mathbf{S}_g\mathbf{T}^\top.
\label{eq:prior}
\end{equation}
The global branch transfers these corrections using similarities between global descriptors~\cite{proker}. The following kernel gives higher similarity to descriptors with more closely aligned directions. For unit vectors, the bandwidth $\beta$ controls how rapidly similarity decreases with angular separation:
\begin{equation}
k_\beta(\mathbf{a},\mathbf{b})=\exp[-\beta(1-\mathbf{a}^\top\mathbf{b})],\qquad \beta>0.
\label{eq:basekernel}
\end{equation}
Throughout, $x$ and $z$ denote arbitrary image views; the arguments of a kernel determine whether each view comes from the support or query collection. Let $K_g(x,z)=k_{\beta_g}(\mathbf{g}(x),\mathbf{g}(z))$. Evaluating a kernel on two ordered collections gives the matrix of all pairwise similarities. The matrix $\mathbf{G}_g=K_g(\mathcal{S},\mathcal{S})$ records similarities among support entries, while $\mathbf{H}_g=K_g(\mathcal{Q},\mathcal{S})$ links queries to support entries. We solve a linear system for the coefficient matrix $\boldsymbol{\alpha}_g$, which has one coefficient per support entry and class. The matrix $\mathbf{G}_g$ accounts for relationships among the support entries, so their coefficients are determined jointly. Multiplication by $\mathbf{H}_g$ transfers the fitted corrections to each query. With regularization parameter $\lambda>0$ and identity matrix $\mathbf{I}_N\in\mathbb{R}^{N\times N}$,
\begin{equation}
\boldsymbol{\alpha}_g=(\mathbf{I}_N+\mathbf{G}_g/\lambda)^{-1}\mathbf{R}_{\mathbf{T}},\qquad
\mathbf{F}_{\mathrm{global}}(\mathcal{Q})=\mathbf{F}_0(\mathcal{Q})+\mathbf{H}_g\boldsymbol{\alpha}_g.
\label{eq:globalprediction}
\end{equation}
Thus $\mathbf{H}_g\boldsymbol{\alpha}_g$ supplies the class-score adjustment added to the CLIP prior. The solve has an equivalent quadratic characterization: it keeps the coefficients close to the support residual target while penalizing the squared kernel-space norm of the correction. This penalty couples the coefficients through $\mathbf{G}_g$, yielding the linear system in Eq.~\eqref{eq:globalprediction}. Since $(\mathbf{I}_N+\mathbf{G}_g/\lambda)^{-1}=\lambda(\mathbf{G}_g+\lambda \mathbf{I}_N)^{-1}$, $\lambda$ controls both regularization and correction amplitude. All branches use this parameterization with the unscaled CLIP prior.

\subsection{Extracting native-layer regional descriptors}
\label{sec:regional-features}

Intermediate feature grids retain regional information from different stages of the encoder. A single encoder pass per view yields $J=4$ regional descriptors at each of three selected layers, indexed by $\ell\in\{1,2,3\}$. A fixed $2\times2$ pooling grid summarizes the content of four spatial regions. The pooling operation follows the structure of each backbone: ViT averages normalized token directions, while RN50 retains activation strength during pooling. Each descriptor $\mathbf{v}_{\ell,a}(x)$ retains the original channel width $D_\ell$ of its layer.

\paragraph{ViT-B/16}
Branches $\ell=1,2,3$ use blocks 4, 8, and 12, respectively. For input view $x$, $\mathbf{t}_{\ell,u}(x)\in\mathbb{R}^{D_\ell}$ denotes the patch token at spatial position $u$ in the selected block. At each layer, $\mathcal{I}_a$ contains the token positions in spatial bin $a\in\{1,\ldots,J\}$, and $|\mathcal{I}_a|$ is the number of positions in that bin. We apply parameter-free LayerNorm, denoted by $\mathrm{LN}$, across each token's channels, followed by L2 normalization and regional averaging:
\begin{equation}
\mathbf{v}_{\ell,a}(x)=\Normalize\!\left(\frac{1}{|\mathcal{I}_a|}\sum_{u\in \mathcal{I}_a}\Normalize(\mathrm{LN}(\mathbf{t}_{\ell,u}(x)))\right),
\label{eq:vitregion}
\end{equation}
Each $\Normalize$ operation divides its input vector by its L2 norm. The inner operation gives each token direction equal weight before averaging; the outer operation makes the resulting regional descriptor a unit vector. Thus Eq.~\eqref{eq:vitregion} first normalizes individual tokens, averages them within each region, and then normalizes the regional average. All four regional vectors are retained, together with the projected CLS descriptor used by the global branch.

\paragraph{RN50}
Branches $\ell=1,2,3$ use C3, C4, and C5, respectively. These stages provide nonnegative post-ReLU feature maps. Generalized-mean (GeM) pooling~\cite{gem} aggregates each of the four bins in an adaptive $2\times2$ grid:
\begin{equation}
u_{\ell,a,d}(x)=\left(\frac{1}{|\mathcal{I}_a|}\sum_{u\in \mathcal{I}_a}h_{\ell,d,u}(x)^3\right)^{1/3},\qquad
\mathbf{v}_{\ell,a}(x)=\Normalize(\mathbf{u}_{\ell,a}(x)).
\label{eq:rnregion}
\end{equation}
Here $h_{\ell,d,u}$ is the scalar activation of channel $d\in\{1,\ldots,D_\ell\}$ at position $u$, and $\mathbf{u}_{\ell,a}$ stacks the pooled channel values $u_{\ell,a,d}$. The fixed exponent of three emphasizes strong responses. Normalization is applied after pooling so that activation magnitudes contribute to the regional summary. RN50 batch-normalization (BatchNorm) statistics remain frozen. The global branch uses the original attention-pooled feature.

For a single query $q$, the regional descriptor matrix $\mathbf{Q}_\ell(q)\in\mathbb{R}^{J\times D_\ell}$ has row $a$ equal to $\mathbf{v}_{\ell,a}(q)^\top$. Applying the same notation to each support view $s_i$, the support tensor $\mathbf{S}_\ell\in\mathbb{R}^{N\times J\times D_\ell}$ has $i$th slice $\mathbf{Q}_\ell(s_i)$. Thus $\mathbf{Q}_\ell(q)$ describes one image, while $\mathcal{Q}$ denotes the collection of queries.

\subsection{Calibrating regional kernels}
\label{sec:regional-kernels}

Augmentation can move the same visual content into different regions of two image views. To match this content across positions, we compare every region in one image with every region in the other and average their kernel similarities. We then normalize this average by the images' self-similarities so that the resulting set embeddings have a common norm. At each depth, the shared bandwidth is $\beta_f=r\beta_g$, where $r$ is the regional-to-global bandwidth ratio. Define
\begin{equation}
\begin{aligned}
A_\ell(x,z)&=\frac{1}{J^2}\sum_{a=1}^{J}\sum_{a'=1}^{J}
 k_{\beta_f}(\mathbf{v}_{\ell,a}(x),\mathbf{v}_{\ell,a'}(z)),\\
K_\ell(x,z)&=\frac{A_\ell(x,z)}{\sqrt{A_\ell(x,x)A_\ell(z,z)}}.
\end{aligned}
\label{eq:setkernel}
\end{equation}
The average $A_\ell$ is a standard kernel-mean construction~\cite{kme}: each image is represented by the mean of its regional embeddings in kernel space. This comparison captures similarity between sets of regional descriptors and is invariant to their ordering. Dividing by the geometric mean of self-similarities removes variation in set-embedding norms and gives $K_\ell(x,x)=1$.

We next measure similarity relative to a common reference defined by the support bank. Let $\boldsymbol{\psi}_\ell(x)$ be the unit-norm kernel-space embedding associated with $K_\ell$, so that $K_\ell(x,z)=\langle\boldsymbol{\psi}_\ell(x),\boldsymbol{\psi}_\ell(z)\rangle$. Its support mean is $\bar{\boldsymbol{\psi}}_\ell=N^{-1}\sum_i\boldsymbol{\psi}_\ell(s_i)$. Centering subtracts this same mean from every support and query embedding. The required statistics are the mean similarity $m_\ell(x)$ between a view and the support bank, and the mean similarity $b_\ell$ among support entries:
\begin{equation}
 m_\ell(x)=\frac{1}{N}\sum_i K_\ell(x,s_i),\qquad
 b_\ell=\frac{1}{N^2}\sum_{i,j}K_\ell(s_i,s_j).
\label{eq:supportmean}
\end{equation}
The same definition gives $m_\ell(z)$ for the other view. These statistics allow centered inner products to be computed directly from kernel values:
\[
\begin{aligned}
&\langle\boldsymbol{\psi}_\ell(x)-\bar{\boldsymbol{\psi}}_\ell,
\boldsymbol{\psi}_\ell(z)-\bar{\boldsymbol{\psi}}_\ell\rangle\\
&\qquad=K_\ell(x,z)-m_\ell(x)-m_\ell(z)+b_\ell.
\end{aligned}
\]
Thus both views are compared relative to the support mean in kernel space. Since $K_\ell(x,x)=1$, the squared length after centering is $1-2m_\ell(x)+b_\ell$. We then normalize the centered embeddings to control variation in their lengths. With $d_\ell^+(x)=\max\{1-2m_\ell(x)+b_\ell,10^{-10}\}$,
\begin{equation}
\widehat K_\ell(x,z)=
\frac{K_\ell(x,z)-m_\ell(x)-m_\ell(z)+b_\ell}
{\sqrt{d_\ell^+(x)d_\ell^+(z)}}.
\label{eq:calibrated}
\end{equation}
The numerator measures similarity between the centered embeddings; the denominator normalizes their magnitudes, with a floor to ensure numerical stability. Support and query images use the same support reference, so each query is calibrated independently of other queries in its batch. The global branch retains its uncentered kernel.

\subsection{Fitting and selecting the residual mixture}
\label{sec:selective-integration}

Different layers can assign different similarities to the same pair of images. Each branch therefore provides a distinct way to transfer the shared support residual to queries. Let $\mathcal B=\{g,1,2,3\}$ index all branches, with $\mathcal K_g=K_g$ and $\mathcal K_\ell=\widehat K_\ell$ for $\ell=1,2,3$. This notation covers both the uncentered global kernel and the calibrated regional kernels. Each branch $b\in\mathcal B$ uses its own kernel matrices and is fitted independently:
\begin{equation}
\mathbf{G}_b=\mathcal K_b(\mathcal{S},\mathcal{S}),\quad \mathbf{H}_b=\mathcal K_b(\mathcal{Q},\mathcal{S}),\quad
\boldsymbol{\alpha}_b=(\mathbf{I}_N+\mathbf{G}_b/\lambda)^{-1}\mathbf{R}_{\mathbf{T}}.
\label{eq:regionsolve}
\end{equation}
Each branch fits the same $\mathbf{R}_{\mathbf{T}}$ through its own $\mathbf{G}_b$, then produces the correction $\boldsymbol{\Delta}_b(\mathcal{Q})=\mathbf{H}_b\boldsymbol{\alpha}_b\in\mathbb{R}^{M\times C}$. For a single query, write $\mathbf{H}_b(q)=\mathcal K_b(q,\mathcal{S})\in\mathbb{R}^{1\times N}$ and $\boldsymbol{\Delta}_b(q)=\mathbf{H}_b(q)\boldsymbol{\alpha}_b\in\mathbb{R}^{1\times C}$. Different depths therefore use their own image relationships to predict corrections to the same class scores. Independent fitting preserves these distinct relationships, while the common $C$-dimensional output allows the predictions to be fused with each layer's native feature dimension retained. The weight vector $\mathbf{w}=(w_1,w_2,w_3)^\top$ sets the relative contributions of the regional branches, and $\rho$ balances their combined correction against the global correction. For $w_\ell\geq0$, $\sum_{\ell=1}^3w_\ell=1$, and $\rho\in[0,1]$, predict
\begin{equation}
\mathbf{F}_{\mathbf{T}}(\mathcal{Q})=\mathbf{F}_0(\mathcal{Q})+(1-\rho)\mathbf{H}_g\boldsymbol{\alpha}_g+
\rho\sum_{\ell=1}^{3}w_\ell \mathbf{H}_\ell\boldsymbol{\alpha}_\ell.
\label{eq:prism}
\end{equation}
The single-query prediction $\mathbf{F}_{\mathbf{T}}(q)\in\mathbb{R}^{1\times C}$ is the corresponding row of $\mathbf{F}_{\mathbf{T}}(\mathcal{Q})$. Classification assigns the class with the largest score. The predictor starts with the CLIP prior and adds the weighted global and regional corrections. The independently fitted predictions are reused across candidate mixture weights. The setting $\rho=0$ recovers the global predictor.

Validation first selects the global bandwidth and regularization $\lambda$, then
the regional bandwidth and mixture weights. All regional branches use the
selected $\lambda$. These parameters remain fixed during subsequent prototype
adaptation and evaluation.

\subsection{Prototype adaptation with fixed kernels}
\label{sec:prototype-adaptation}

Each row of $\mathbf{T}$ is a text-derived representative vector, or prototype, for one class. LAMP-F adapts these vectors to the target task using the labeled support images. A prototype update changes both the initial query scores and the support scores used to form the residual target. We account for both effects while keeping the selected visual kernels and mixture weights fixed. The trainable matrix $\mathbf{W}\in\mathbb{R}^{C\times d_g}$ is initialized to $\mathbf{T}$:
\begin{equation}
\boldsymbol{\Gamma}(\mathbf{W})=\Normalize(\mathbf{W})-\Normalize(\mathbf{T}),\qquad \mathbf{P}(\mathbf{W})=\mathbf{T}+\boldsymbol{\Gamma}(\mathbf{W}),
\label{eq:prototype}
\end{equation}
Here $\boldsymbol{\Gamma}(\mathbf{W})$ is the change in class directions after row-wise normalization, and $\mathbf{P}(\mathbf{W})$ is the adapted prototype matrix. With unit rows in $\mathbf{T}$, $\mathbf{P}(\mathbf{W})$ equals $\Normalize(\mathbf{W})$ in exact arithmetic. Writing the update as a difference also makes $\boldsymbol{\Gamma}(\mathbf{T})=\mathbf{0}$, so initialization recovers LAMP exactly. Prototype training uses one mean descriptor per support image, $\bar{\mathbf{g}}_i=V^{-1}\sum_{v=1}^{V}\mathbf{g}(x_{i,v})$. This mean is used directly, so its linear class scores equal the average scores across views. Following CLAP~\cite{clap}, we optimize
\begin{equation}
\mathcal{L}(\mathbf{W})=\frac{1}{CK}\sum_{i=1}^{CK}\mathrm{CE}(100\mathbf{P}(\mathbf{W})\bar{\mathbf{g}}_i,y_i)
+\frac{1}{C}\sum_{c=1}^{C}a_c\|\mathbf{W}_c-\mathbf{T}_c\|_2^2,
\label{eq:trainingloss}
\end{equation}
Here $\mathrm{CE}$ is cross-entropy, and $\mathbf{W}_c,\mathbf{T}_c\in\mathbb{R}^{1\times d_g}$ are the $c$th rows of their respective matrices. The first term adapts the prior scores to the support labels; the second anchors each class parameter to its original text vector. The fixed weight $a_c$ is the mean softmax confidence assigned to the correct class across the original views of class $c$, computed from unscaled CLIP scores. A larger weight anchors the class prototype more strongly to its text vector.

AdamW updates the prototypes using the support-set classification loss and text-prototype anchor in Eq.~\eqref{eq:trainingloss}. Checkpoint selection evaluates the complete residual predictor on labeled validation data. The cached response derived below accelerates these repeated full-predictor evaluations as the prototypes change. The factor 100 scales training logits. The kernel bank continues to use ten separate views per image.

The prototype change adds $\mathbf{Q}_g\boldsymbol{\Gamma}(\mathbf{W})^\top$ to the query prior and changes the support residual to $\mathbf{R}_{\mathbf{T}}-\mathbf{S}_g\boldsymbol{\Gamma}(\mathbf{W})^\top$. An improvement in the support prior therefore changes the correction that the kernel branches need to supply. For $b\in\mathcal B$, let $\mathbf{L}_b=(\mathbf{I}_N+\mathbf{G}_b/\lambda)^{-1}$ and write $\boldsymbol{\alpha}_b(\mathbf{P})$ for the coefficients associated with an adapted prototype matrix $\mathbf{P}=\mathbf{P}(\mathbf{W})$, with $\boldsymbol{\alpha}_b(\mathbf{T})=\boldsymbol{\alpha}_b$. Linearity in the residual gives
\[
\begin{aligned}
\boldsymbol{\alpha}_b(\mathbf{P})
&=\mathbf{L}_b[\mathbf{R}_{\mathbf{T}}-\mathbf{S}_g\boldsymbol{\Gamma}(\mathbf{W})^\top]\\
&=\boldsymbol{\alpha}_b(\mathbf{T})-
\underbrace{\mathbf{L}_b\mathbf{S}_g}_{\mathbf{B}_b}
\boldsymbol{\Gamma}(\mathbf{W})^\top.
\end{aligned}
\]
The fixed matrix $\mathbf{B}_b$ therefore converts a prototype change into a coefficient update, allowing repeated updates to reuse the same linear-system response. With $\pi_g=1-\rho$ and $\pi_\ell=\rho w_\ell$, we precompute the support responses and combine their effects on a query collection:
\begin{equation}
\mathbf{B}_b=\mathbf{L}_b\mathbf{S}_g,\qquad \mathbf{E}(\mathcal{Q})=\sum_{b\in\mathcal{B}}\pi_b\mathbf{H}_b\mathbf{B}_b
\label{eq:response}
\end{equation}
The matrix $\mathbf{B}_b\in\mathbb{R}^{N\times d_g}$ is obtained by solving the linear system for branch $b$ with the global support descriptors as its right-hand side. It is shared across query collections. Multiplying it by $\mathbf{H}_b$ transfers the response to the queries, and the weighted sum gives $\mathbf{E}(\mathcal{Q})\in\mathbb{R}^{M\times d_g}$. This query-specific response can be reused across prototype checkpoints for the same collection. The adapted prediction is then exactly
\begin{equation}
\mathbf{F}_{\mathbf{P}(\mathbf{W})}(\mathcal{Q})=\mathbf{F}_{\mathbf{T}}(\mathcal{Q})+[\mathbf{Q}_g-\mathbf{E}(\mathcal{Q})]\boldsymbol{\Gamma}(\mathbf{W})^\top.
\label{eq:analyticupdate}
\end{equation}
The term $\mathbf{Q}_g\boldsymbol{\Gamma}(\mathbf{W})^\top$ updates the prior, while $-\mathbf{E}(\mathcal{Q})\boldsymbol{\Gamma}(\mathbf{W})^\top$ updates the fitted corrections for the changed support residual. Together, these terms account for both effects of prototype adaptation. Once the response is cached, evaluating a new prototype matrix only requires matrix products with $\boldsymbol{\Gamma}(\mathbf{W})^\top$. LAMP-F optimizes only $Cd_g$ prototype parameters; the encoders, regional descriptors, and fusion weights remain frozen.

%% file: experiments.tex
\section{Experiments}

We evaluate recognition accuracy at different support budgets and examine how
class-prototype adaptation and native-layer integration contribute to performance.
We also test transfer to new domains without fitting on target data and measure
deployment costs. Comparisons with published methods establish overall performance,
while paired controls assess individual design choices.

\subsection{Evaluation protocol}
\label{sec:evaluation-protocol}
We evaluate training-free LAMP and its prototype-adapted extension, LAMP-F, with CLIP encoders from the ViT and ResNet families. The main comparison covers eleven datasets using ViT-B/16 and RN50, with 1, 2, 4, 8, and 16 support images per class and three support-sampling seeds. Additional experiments use RN101 and ViT-B/32. We report top-1 accuracy and give each dataset equal weight when computing the mean.

We follow the established benchmark partitions and evaluation conventions used in previous studies~\cite{coop,tipadapter,proker}. Labeled validation images are not counted toward the support budget. Paired controls use the same support images, augmented views, prompts, and query images. Results taken from published baselines retain their original protocols. Confidence intervals for our experiments are conditional on the selected models.

Ten datasets have distinct validation and test lists. For ImageNet, support
images come from the training partition, while the same official 50,000-image
validation partition is used for both parameter selection and benchmark
scoring, following the released Tip-Adapter and ProKeR implementations.
The eleven-dataset aggregate therefore combines these evaluation protocols.
Published comparisons retain their different selection procedures:
CLIP-LoRA shares hyperparameters across tasks, whereas LAMP uses
dataset-specific validation.

Paired analyses of the three-seed benchmark and its controls use 5,000
paired crossed bootstrap replicates, with shared support-seed and
class-stratified query-identity draws. Selected models remain fixed during
resampling. The benchmark tasks were used during method development;
these intervals do not cover repeated-selection or development uncertainty.

\input{main_published_comparison}

\subsection{Paired gains over the global baseline}
\label{sec:heldout-results}
Our paired global baseline reproduces ProKeR's residual predictor in
Eq.~\eqref{eq:globalprediction}, using the final descriptor from frozen CLIP.
Table~\ref{tab:reviewer-eleven-datasets} compares this predictor with LAMP and
LAMP-F across all eleven datasets. The three methods share support banks,
prompts, and selected global kernel parameters, so the comparison measures
the gains from regional integration and the subsequent prototype adaptation.
At 16-shot, the global baseline reaches 82.30\% with ViT-B/16 and 76.62\%
with RN50. LAMP raises these means to 83.30\% and 78.73\%, and LAMP-F to
83.57\% and 79.10\%, respectively.

Averaged equally across datasets and the five support budgets, LAMP improves
accuracy by 1.14 percentage points with ViT and 1.59 with RN50; LAMP-F improves
it by 1.96 and 2.54 points. Both methods improve mean accuracy at every
support budget on both backbones.
Section~\ref{sec:dataset-native-effects} examines the gains by dataset.

\input{reviewer_eleven_dataset_table}

\subsection{Dataset-wise performance gains}
\label{sec:dataset-native-effects}
At four shots, integrating native-layer features improves mean accuracy across the eleven datasets by 1.446 points with ViT and 1.568 with RN50 (Table~\ref{tab:eleven-dataset-effects}). The largest gains occur on EuroSAT: 5.860 and 6.695 points, respectively. Mean accuracy increases on all eleven datasets with ViT and on ten with RN50; the RN50 change on SUN397 is $-0.032$ points. Native-layer integration is particularly effective for satellite imagery and, with ViT, texture recognition.

Prototype adaptation benefits a different set of tasks. Its largest 4-shot gains occur on SUN397, at 5.620 points with ViT and 6.715 with RN50. Prototype adaptation and native-layer integration thus benefit different tasks: the former adjusts class directions, while the latter changes how support evidence is used to correct the initial CLIP scores.

\input{v18_eleven_dataset_effects_table}

\subsection{Layer depth, pooling, and calibration}
\label{sec:v4-native-pooling}
\label{sec:v4-simplification}
We examine the source of LAMP's gains through three groups of controls:
the information used by the kernel branches, the representation retained
at each depth, and the calibration and weighting of their corrections.
All comparisons use paired support banks at 4-shot.

\paragraph{Information from different depths}
MULTIBAND constructs three kernels from the same final CLIP global descriptor,
using bandwidths $0.5\beta_g$, $\beta_g$, and $2\beta_g$. Each calibrated branch
fits the shared support residual independently, and their corrections are
mixed with the original global predictor. This control tests the benefit of
changing the similarity scale while keeping the visual representation fixed.
Across the ten non-ImageNet datasets, native-region kernels outperform
MULTIBAND by 1.286 points with ViT and 1.635 with RN50 under matched
113-candidate mixture searches (Table~\ref{tab:mechanism-summary}, top).
The family-adjusted 98.75\% intervals are positive for both backbones,
supporting the value of information from
multiple encoder depths.

\paragraph{Depth and spatial pooling}
We compare four representations on Pets, Food101, DTD, and EuroSAT.
REGION retains four regional descriptors at each of three depths.
CENTROID averages the four normalized descriptors at each depth and normalizes
the result; WHOLE pools each feature map directly into one descriptor.
FINAL uses only the deepest native feature map, with three bandwidths.
Thus WHOLE versus FINAL tests the use of multiple depths, while REGION versus
CENTROID or WHOLE tests the effect of retaining separate regional descriptors.
All four families evaluate 113 mixtures; REGION, CENTROID, and WHOLE use
fixed regional bandwidth ratio $r=1$.

\begin{table}[!tbp]
\centering
\renewcommand{\baselinestretch}{1}\fontsize{9}{11}\selectfont
\setlength{\tabcolsep}{4pt}
\renewcommand{\arraystretch}{1.08}
\caption{Representation controls at 4-shot with 113 mixture candidates per family.
Top: mean gains in percentage points over ten non-ImageNet datasets.
Bottom: mean accuracy (\%) over Pets, Food101, DTD, and EuroSAT, averaged
equally across datasets and three support seeds. MULTIBAND uses the final
CLIP global descriptor; FINAL uses the deepest native feature map.}
\label{tab:mechanism-summary}
\begin{tabular*}{\linewidth}{@{\extracolsep{\fill}}lrr@{}}
\toprule
Comparison or representation & ViT-B/16 & RN50 \\
\midrule
\multicolumn{3}{l}{\textit{Native-layer versus global information (gain)}} \\
REGION $-$ MULTIBAND & +1.286 & +1.635 \\
\midrule
\multicolumn{3}{l}{\textit{Depth and pooling (accuracy)}} \\
REGION: four descriptors per depth & 82.993 & 77.060 \\
CENTROID: mean of regional descriptors & 82.991 & 77.067 \\
WHOLE: one descriptor per depth & 83.021 & 76.988 \\
FINAL: deepest native map only & 81.828 & 76.083 \\
\bottomrule
\end{tabular*}
\end{table}

WHOLE exceeds FINAL by 1.193 [0.648, 1.806] points with ViT and
0.905 [0.499, 1.384] with RN50, with pointwise 95\% intervals in brackets.
Combining depths therefore improves accuracy even when each map is pooled
into a single descriptor. REGION and WHOLE differ by $-0.028$
[$-0.162$, 0.129] and 0.072 [$-0.083$, 0.220] points, respectively.
The similar mean accuracies of REGION, CENTROID, and WHOLE indicate that
compact descriptors retain much of the observed gain from native-layer
information.

\paragraph{Calibration and layer weighting}
We next restore bandwidth selection and simplify the full LAMP model on
the same four datasets. S1 replaces regional sets with whole-map descriptors;
S2 additionally removes support centering and diagonal normalization;
S3 further fixes the three layer weights to $1/3$.
LAMP, S1, and S2 each search 337 configurations, while S3 searches 22.
S1 thus uses the WHOLE representation with a larger bandwidth search.

The clearest calibration effect occurs with RN50: removing both operations
reduces mean accuracy by 0.627 points, with an adjusted 99.1667\% interval of
[$-0.979$, $-0.277$] for S2 minus S1. On EuroSAT, accuracy falls from
82.802\% to 80.811\%. The ViT change is $+0.215$ [$-0.075$, $+0.514$],
showing a less consistent benefit from calibration on this backbone.
Replacing regional sets with whole-map descriptors also has task-dependent
effects: ViT accuracy on DTD falls from 66.923\% to 66.056\%, while Pets
and Food101 improve slightly.

Equal layer weights reduce the means by a further 0.331 points with ViT and
0.180 with RN50. Their adjusted intervals include zero; this comparison
assesses the complete selected-weight and equal-weight strategies, including
their search budgets. Even after all three simplifications, S3 exceeds the
paired global baseline by 2.179 and 1.117 points. These results locate the
main gain in native-layer information and the strongest calibration benefit
in RN50.

\subsection{Performance gains of LAMP-F over LAMP}
\label{sec:prototype-dataset-analysis}
\label{sec:selection-analysis}
This experiment isolates prototype adaptation by retaining LAMP's selected
kernels and mixture weights. Across eleven datasets and five support budgets,
LAMP-F improves the three-seed mean in 53 of 55 settings with ViT and
52 of 55 with RN50. With ViT, there is one tie (Pets, 16-shot) and one
decrease (Caltech101, 8-shot; $-0.068$ points); the three RN50 ties all occur
on Pets, at 1, 8, and 16 shots. Mean gains decrease from 1.097 and 1.463 points
at 4-shot to 0.274 and 0.366 at 16-shot for ViT and RN50, respectively.
SUN397 benefits most, with gains concentrated at 1--4 shots.

\paragraph{Comparison with an independently adapted global predictor}
We also train a separate ProKeR+CLAP model~\cite{proker,clap} on the ten
non-ImageNet datasets at 4-shot. It shares the support data, prompts,
validation labels, and checkpoint-search budget with LAMP-F. Both methods
evaluate 33 learning-rate/checkpoint candidates; LAMP-F additionally inherits the
337-candidate regional-mixture selection.
LAMP-F improves mean accuracy by 1.840 points with ViT and 1.949 with RN50;
the family-adjusted intervals are [1.451, 2.283] and [1.483, 2.417],
respectively. This comparison assesses the complete LAMP-F adaptation
procedure against a global predictor with independently optimized prototypes.

\subsection{Support-draw robustness and validation-label budget}
\label{sec:validation-budget}
\begin{figure}[!tbp]
\centering\includegraphics[width=\linewidth]{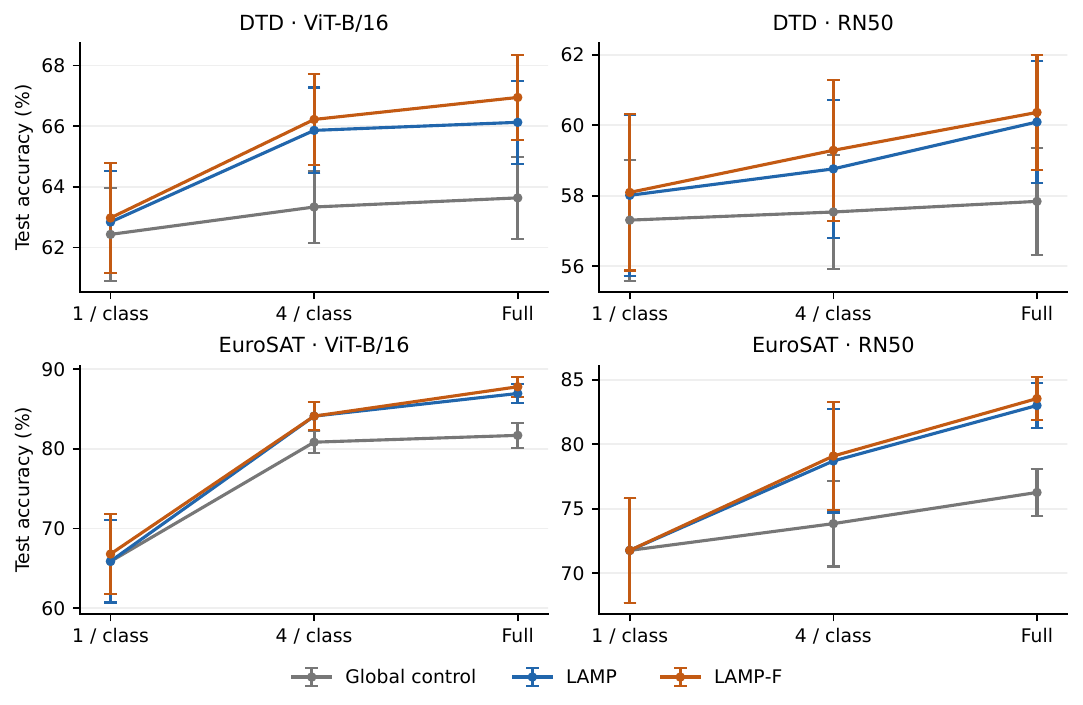}
\caption{Effect of the validation-label budget at four shots on DTD and EuroSAT.
Points show mean test accuracy over ten new support draws, and error bars
indicate one sample SD. The full validation split contains 24 images per class
on DTD and 540 on EuroSAT. Methods and support draws use the same nested
validation subsets; the complete parameter search is repeated for each setting.}
\label{fig:validation-budget}
\end{figure}

To assess the annotation requirements of model selection, we sample ten new
4-shot support sets on DTD and EuroSAT. For each backbone and support set,
we repeat the complete parameter search using one validation image per class,
four images per class, or the full validation split. The nested validation
subsets are fixed across support draws and backbones. Confidence intervals
for this study use paired Student-$t$ statistics over the ten support draws,
conditional on the fixed validation subset and test set.
Figure~\ref{fig:validation-budget} summarizes all 120 configurations.
With the full validation split, LAMP improves on the global baseline by
2.494 and 2.252 points on DTD with ViT and RN50, respectively; the corresponding
gains on EuroSAT are 5.253 and 6.738 points. All four adjusted confidence
intervals lie above zero, showing that the gains persist across additional
support draws.

The validation budget strongly affects model selection. With one validation
image per class on EuroSAT, the procedure selects the global-only candidate
in 19 of 20 combinations of backbone and support set. Increasing the budget
to four images per class yields mean gains of 3.267 and 4.857 points with ViT
and RN50, respectively. Labeled validation data therefore play an important
role in selecting effective native-layer corrections.

\subsection{Additional backbones and frozen-target transfer}
\label{sec:additional-backbones}
\label{sec:frozen-transfer}
To examine adaptation across encoder architectures, depths, and patch sizes,
we compare training-free methods on ImageNet at 16 shots using four CLIP
backbones (Table~\ref{tab:imagenet-published}). LAMP improves over the published ProKeR
scores by 1.28, 1.13, 1.00, and 0.85 percentage points for RN50, RN101,
ViT-B/32, and ViT-B/16, respectively. LAMP achieves the highest accuracy
among the methods listed in this table on each backbone.

\input{imagenet_backbone_table}

Paired comparisons assess the contribution of native-layer corrections with
RN101 and ViT-B/32. With RN101, LAMP improves mean accuracy across the ten
non-ImageNet datasets at every support budget by 1.196--2.333 points; all pointwise 95\%
confidence intervals lie above zero (Table~\ref{tab:rn101-transfer}). Across
all eleven datasets and five support budgets, five of the 55 changes are
negative, including Pets at 2-shot
($-0.645$ [$-1.663$, $-0.018$]). On ImageNet at 16-shot, LAMP achieves 68.523\%
with RN101 and 69.125\% with ViT-B/32, with the latter exceeding its paired
baseline by 0.465 points. These results demonstrate the use of the same
training-free residual formulation with different encoder depths and patch sizes.

\input{rn101_main_table}

To evaluate robustness to distribution shift, we apply the ViT-B/16 and RN50
models fitted on ImageNet at 16-shot to ImageNet-V2 and
ImageNet-Sketch~\cite{imagenetv2,imagenetsketch}. All fitted components remain
fixed, with no further fitting or parameter selection on either target dataset.

\begin{table}[tbp]
\centering
\renewcommand{\baselinestretch}{1}\fontsize{9}{11}\selectfont
\setlength{\tabcolsep}{2pt}
\caption{Transfer of frozen 16-shot models: mean top-1 accuracy (\%) over three seeds
and LAMP's paired gain over the global-feature-only baseline. Gains are in
percentage points, with pointwise 95\% confidence intervals from 5,000 paired
bootstrap draws.}
\label{tab:frozen-transfer}
\begin{tabular*}{\linewidth}{@{\extracolsep{\fill}}llrrl@{}}\toprule
Backbone & Target & Global-only & LAMP & Gain [95\% interval] \\\midrule
ViT-B/16 & ImageNet-V2 & 65.843 & 66.297 & +0.453 [+0.110, +0.837] \\
ViT-B/16 & ImageNet-Sketch & 49.258 & 49.910 & +0.652 [+0.363, +0.914] \\
RN50 & ImageNet-V2 & 56.220 & 57.077 & +0.857 [+0.537, +1.177] \\
RN50 & ImageNet-Sketch & 35.771 & 36.631 & +0.860 [+0.689, +1.059] \\
\bottomrule\end{tabular*}\end{table}

LAMP improves accuracy in all twelve combinations of seed, target dataset,
and backbone. The pointwise 95\% confidence intervals for all four aggregate
gains in Table~\ref{tab:frozen-transfer} lie above zero, showing that native-layer
corrections remain useful under both distribution shifts. With ViT, LAMP
achieves 66.297\% on ImageNet-V2 and 49.910\% on ImageNet-Sketch, compared
with 74.105\% on the source ImageNet benchmark. 

\subsection{Hyperparameter behavior and layer contributions}
\label{sec:hyperparameters}
\begin{figure}[!tbp]
\centering\includegraphics[width=\linewidth]{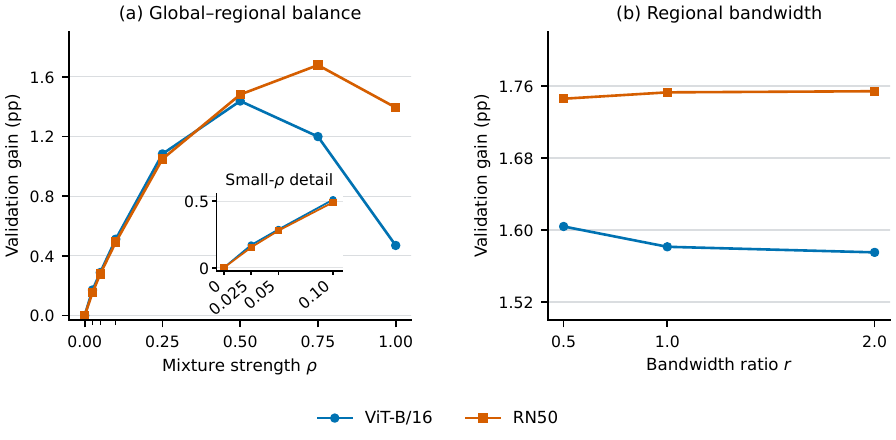}
\caption{Regional parameter profiles at 4-shot, averaged equally over eleven
datasets and three support seeds. Each point is the validation gain over
the global baseline after reselecting the other regional parameters.
Left: mixture strength $\rho$ in Eq.~\eqref{eq:prism}, with an inset showing
the small-$\rho$ values. Right: bandwidth ratio $r=\beta_f/\beta_g$
for the regional kernel in Eq.~\eqref{eq:setkernel}.}
\label{fig:hyperparameters}
\end{figure}

We examine how the parameters in the residual mixture affect validation
performance. The mixture strength $\rho$ and layer weights $\mathbf w$ appear
in Eq.~\eqref{eq:prism}; the ratio $r=\beta_f/\beta_g$ sets the regional
bandwidth in Eq.~\eqref{eq:setkernel}, using the kernel in Eq.~\eqref{eq:basekernel}.
Figure~\ref{fig:hyperparameters} summarizes 4-shot results over eleven datasets
and three support-sampling seeds. At each plotted value, the remaining
regional parameters are reselected for each support bank.

\paragraph{Global--regional balance}
The mean validation profile peaks at $\rho=0.5$ for ViT-B/16 and $\rho=0.75$
for RN50, with gains of 1.437 and 1.678 points over the global baseline.
At $\rho=1$, the gains are 0.470 and 1.394 points.
Combining the global and native-layer corrections therefore gives the
highest mean validation accuracy among the tested mixture strengths.

\paragraph{Regional bandwidth}
For $r\in\{0.5,1,2\}$, the mean validation gains range from 1.575 to 1.604
points with ViT and from 1.746 to 1.754 with RN50. Selecting the mixture
separately at each ratio yields similar mean performance across these
three similarity scales.

\input{selected_weights_figure}
\paragraph{Contributions of different depths}
Figure~\ref{fig:effective-weights} displays the four coefficients in
Eq.~\eqref{eq:prism}: $(1-\rho,\rho w_1,\rho w_2,\rho w_3)$.
The factor $\rho$ determines the total regional contribution, and
$\mathbf w$ distributes it among the three depths. At 4-shot, the selected
mixtures favor the global branch on Pets and assign a larger regional
contribution on DTD and EuroSAT than on Pets. With ViT, Flowers102 places its
largest regional
weight on the late layer, while DTD favors the middle layer. These patterns
show how the selected mixture varies with the task.
Section~\ref{sec:v4-native-pooling} measures the corresponding accuracy effects
of combining depths.

\subsection{Computational cost}
\label{sec:computational-cost}
\input{deployment_cost_main_table}
Table~\ref{tab:deployment-cost} measures image-to-label inference on one
RTX~4090 at 16 shots. On ImageNet with ViT-B/16, LAMP increases single-image
latency from 14.22 to 64.83\,ms, while batch-128 throughput changes from
113.38 to 107.89 images/s. With RN50, the corresponding values are
14.81 to 65.86\,ms and 118.64 to 112.28 images/s. Regional matching therefore
has a larger relative effect on single-image latency than on batched throughput.
Peak allocated GPU memory at batch 128 increases from 2.71 to 7.64\,GiB
with ViT and from 3.02 to 11.55\,GiB with RN50. LAMP-F adds the prototype
response storage and reaches 8.56 and 13.39\,GiB, respectively.

Compact regional representations offer a storage--accuracy trade-off.
On the DTD 4-shot diagnostic bank, CENTROID matches REGION's accuracy on
both backbones using one quarter of the native descriptor storage.

%% file: main_published_comparison.tex
\subsection{Comparison with state-of-the-art methods}
\label{sec:published-comparison}
We compare LAMP and LAMP-F with representative published methods for prompt
learning, feature adaptation, and support matching across five support budgets.
Tables~\ref{tab:published-vit} and~\ref{tab:published-rn-validation} report mean
accuracy over eleven datasets for ViT-B/16 and RN50. The methods included in
each comparison depend on their backbone requirements and the availability of
published results. For example, MaPLe~\cite{maple} inserts deep visual prompts
into Transformer blocks and is evaluated with ViT-B/16. SSP~\cite{ssp} reports
eleven-dataset mean accuracy at all five support budgets for RN50, but does
not report the corresponding means for ViT-B/16. We therefore use the
ViT-B/16 results reported by CLIP-LoRA and CAPEL~\cite{cliplora,chen2026capel}
to compare with methods that require training, and the RN50 results reported
by ProKeR and SSP~\cite{proker,ssp} to compare with training-free methods.

LAMP achieves state-of-the-art mean accuracy among the training-free RN50
methods compared at 4, 8, and 16 shots. At 16 shots, it reaches 78.73\%,
exceeding the best published training-free result by 1.98 percentage points
without gradient optimization.

The ViT-B/16 comparison includes published methods that require training.
LAMP exceeds CLIP-LoRA, the strongest published baseline at 8 and 16 shots,
by 0.35 and 0.30 points, respectively, without gradient-based training.
LAMP-F further raises mean accuracy to 80.94\% and 83.57\%. CLIP-LoRA
updates encoder weights, while LAMP keeps both the encoders and class
prototypes fixed; LAMP-F optimizes only the class prototypes. These results
show that combining residual corrections from multiple depths can achieve
competitive accuracy with a frozen encoder and fixed text prototypes.

\begin{table}[tbp]
\centering
\renewcommand{\baselinestretch}{1}\fontsize{9}{11}\selectfont
\setlength{\tabcolsep}{3pt}
\renewcommand{\arraystretch}{1.08}
\caption{ViT-B/16 comparison: mean top-1 accuracy (\%) over eleven datasets. Published scores are taken from CLIP-LoRA~\cite{cliplora} and CAPEL~\cite{chen2026capel}; results for LAMP and LAMP-F are averaged over three seeds. Boldface and underlining indicate the highest and second-highest means in each column. The numbers in CoOp (4)/(16) denote context length. Venue/year identifies each method's publication; PLOT++ is the enhanced PLOT implementation used in CLIP-LoRA. PR denotes Pattern Recognition, and CVPRW denotes CVPR Workshops. Evaluation protocols are described in Section~\ref{sec:evaluation-protocol}.}
\label{tab:published-vit}
\begin{tabular*}{\linewidth}{@{\extracolsep{\fill}}llrrrrr@{}}
\toprule
Method & Venue/year & 1 & 2 & 4 & 8 & 16 \\
\midrule
CoOp (4)~\cite{coop} & IJCV, 2022 & 67.2 & 70.5 & 73.0 & 76.4 & 79.0 \\
CoOp (16)~\cite{coop} & IJCV, 2022 & 67.6 & 70.8 & 74.0 & 76.8 & 80.0 \\
CoCoOp~\cite{cocoop} & CVPR, 2022 & 68.8 & 70.4 & 71.7 & 73.8 & 75.4 \\
Tip-Adapter-F~\cite{tipadapter} & ECCV, 2022 & 70.9 & 73.7 & 75.6 & 78.3 & 80.7 \\
CLIP-Adapter~\cite{clipadapter} & IJCV, 2024 & 65.7 & 66.8 & 67.7 & 71.5 & 75.5 \\
PLOT++~\cite{plot} & ICLR, 2023 & 70.7 & 74.0 & 76.9 & 79.6 & 82.1 \\
KgCoOp~\cite{kgcoop} & CVPR, 2023 & 69.6 & 71.6 & 73.9 & 76.0 & 77.3 \\
TaskRes~\cite{taskres} & CVPR, 2023 & 70.8 & 72.9 & 74.7 & 78.4 & 80.8 \\
MaPLe~\cite{maple} & CVPR, 2023 & 66.4 & 70.5 & 73.5 & 76.4 & 78.6 \\
ProGrad~\cite{prograd} & ICCV, 2023 & 69.5 & 72.4 & 74.7 & 77.4 & 79.9 \\
CLIP-LoRA~\cite{cliplora} & CVPRW, 2024 & 72.5 & \underline{75.5} & 77.4 & 80.3 & 83.0 \\
CAPEL~\cite{chen2026capel} & PR, 2026 & \underline{72.97} & 74.90 & 77.64 & 80.14 & 82.45 \\
\midrule
\textbf{LAMP} & Ours & 72.18 & 74.27 & \underline{77.70} & \underline{80.65} & \underline{83.30}\\
\textbf{LAMP-F} & Ours & \textbf{73.28} & \textbf{75.62} & \textbf{78.80} & \textbf{80.94} & \textbf{83.57} \\
\bottomrule
\end{tabular*}
\end{table}

On RN50, training-free LAMP outperforms the strongest published method by
0.77, 1.18, and 1.98 points at 4, 8, and 16 shots, respectively.
APE+SSP~\cite{ssp} achieves the highest training-free mean accuracy at 1 and
2 shots, while LAMP's advantage grows with the number of support images.
LAMP-F further improves the means to 65.91, 68.83, 72.65, 75.50, and 79.10\%.
Across the two backbones, native-layer residual integration is competitive
with both training-free support matching and methods that require training,
and prototype optimization provides further gains.
Section~\ref{sec:prototype-dataset-analysis} presents a
paired comparison with independently trained ProKeR+CLAP prototypes.

\begin{table}[tbp]
\centering
\renewcommand{\baselinestretch}{1}\fontsize{9}{11}\selectfont
\setlength{\tabcolsep}{3pt}
\renewcommand{\arraystretch}{1.08}
\caption{RN50 comparison: mean top-1 accuracy (\%) over eleven datasets. Published scores are taken from Table~2 of ProKeR~\cite{proker}, including variants using random Fourier features (RFF), and Table~3 of SSP~\cite{ssp}. Results for LAMP and LAMP-F are averaged over three seeds. The published methods and LAMP are training-free; LAMP-F optimizes class prototypes. Boldface and underlining indicate the highest and second-highest values in each column across all rows. APE+SSP uses the CuPL prompts specified in the source publication. Venue/year identifies the publication reporting each method or RFF variant. Evaluation protocols are described in Section~\ref{sec:evaluation-protocol}.}
\label{tab:published-rn-validation}
\begin{tabular*}{\linewidth}{@{\extracolsep{\fill}}llrrrrr@{}}
\toprule
Method & Venue/year & 1 & 2 & 4 & 8 & 16 \\
\midrule
GDA~\cite{gda} & ICLR, 2024 & 62.19 & 66.19 & 69.77 & 73.30 & 76.04 \\
Tip-Adapter~\cite{tipadapter} & ECCV, 2022 & 62.83 & 64.84 & 66.50 & 68.64 & 70.02 \\
Tip-Adapter with RFF~\cite{tipadapter,proker} & CVPR, 2025 & 62.77 & 64.58 & 66.29 & 68.15 & 69.60 \\
APE~\cite{ape} & ICCV, 2023 & 64.43 & 66.48 & 68.66 & 70.75 & 72.81 \\
APE with RFF~\cite{ape,proker} & CVPR, 2025 & 64.16 & 66.11 & 68.76 & 70.37 & 72.02 \\
APE+SSP~\cite{ssp} & ACM MM, 2024 & \textbf{66.03} & \underline{68.04} & 69.90 & 71.63 & 73.72 \\
ProKeR (RBF)~\cite{proker} & CVPR, 2025 & 64.01 & 67.31 & 70.42 & 73.85 & 76.75 \\
ProKeR with RFF~\cite{proker} & CVPR, 2025 & 64.01 & 67.12 & 70.28 & 73.61 & 76.44 \\
\midrule
\textbf{LAMP} & Ours & 64.79 & 67.50 & \underline{71.19} & \underline{75.03} & \underline{78.73}\\
\textbf{LAMP-F} & Ours & \underline{65.91} & \textbf{68.83} & \textbf{72.65} & \textbf{75.50} & \textbf{79.10} \\
\bottomrule
\end{tabular*}
\end{table}

\paragraph{Results by dataset}
Table~\ref{tab:main-vit16} compares representative published methods at 16
shots on all eleven datasets using ViT-B/16. LAMP-F achieves the highest
accuracy among the listed methods on five datasets: ImageNet, EuroSAT,
Food101, Flowers102, and DTD. It exceeds the best published scores shown for
these datasets by 0.66, 1.58, 0.14, 0.90, and 2.74 points, respectively.
The large gains on texture and satellite-image datasets motivate the
analysis of native-layer features below. The methods have different strengths
across datasets: CLIP-LoRA leads on Aircraft, Cars, and UCF101; CAPEL on
SUN397 and Caltech101; and PLOT++ on Pets.
Table~\ref{tab:imagenet-published} extends the comparison of training-free
methods to ImageNet with four backbones.
\input{main_vit16_comparison}
\FloatBarrier

%% file: main_vit16_comparison.tex
\begin{table}[tbp]
\centering
\renewcommand{\baselinestretch}{1}\fontsize{9}{11}\selectfont
\setlength{\tabcolsep}{3pt}
\renewcommand{\arraystretch}{1.08}
\caption{ViT-B/16 16-shot top-1 accuracy (\%) for representative published methods. Boldface and underlining indicate the highest and second-highest scores in each row. We retain the numerical precision and mean accuracies reported in the source publications. Evaluation protocols are described in Section~\ref{sec:evaluation-protocol}.}
\label{tab:main-vit16}
\begin{tabular*}{\linewidth}{@{\extracolsep{\fill}}lrrrrrr@{}}
\toprule
Dataset & \multicolumn{1}{c}{Tip-Adapter-F} & \multicolumn{1}{c}{PLOT++} & \multicolumn{1}{c}{CLIP-LoRA} & \multicolumn{1}{c}{CAPEL} & \multicolumn{1}{c}{LAMP} & \multicolumn{1}{c}{LAMP-F} \\
 & \multicolumn{1}{c}{\cite{tipadapter}} & \multicolumn{1}{c}{\cite{plot}} & \multicolumn{1}{c}{\cite{cliplora}} & \multicolumn{1}{c}{\cite{chen2026capel}} & & \\
\midrule
ImageNet~\cite{imagenet} & 73.4 & 72.6 & 73.6 & 73.73 & \underline{74.10} & \textbf{74.39} \\
SUN397~\cite{sun397} & 76.0 & 76.0 & 76.1 & \textbf{77.11} & 76.35 & \underline{76.36} \\
FGVC Aircraft~\cite{fgvcaircraft} & 44.6 & 46.7 & \textbf{54.7} & 48.09 & 49.83 & \underline{50.58} \\
EuroSAT~\cite{eurosat} & 85.9 & 92.0 & 92.1 & 88.73 & \underline{93.16} & \textbf{93.68} \\
Stanford Cars~\cite{stanfordcars} & 82.3 & 84.6 & \textbf{86.3} & 84.40 & 84.66 & \underline{84.85} \\
Food101~\cite{food101} & 86.8 & 87.1 & 84.2 & 87.81 & \underline{87.85} & \textbf{87.95} \\
Oxford-IIIT Pets~\cite{oxfordpets} & 92.6 & \textbf{93.6} & 92.4 & \underline{93.57} & 93.38 & 93.38 \\
Flowers102~\cite{flowers102} & 96.2 & 97.6 & 98.0 & 97.97 & \underline{98.85} & \textbf{98.90} \\
Caltech101~\cite{caltech101} & 95.7 & 96.0 & 96.4 & \textbf{96.55} & 96.36 & \underline{96.48} \\
DTD~\cite{dtd} & 70.8 & 71.4 & 72.0 & 73.46 & \underline{75.73} & \textbf{76.20} \\
UCF101~\cite{ucf101} & 83.9 & 85.3 & \textbf{86.7} & 85.51 & 85.96 & \underline{86.50} \\
\midrule
Avg. & 80.7 & 82.1 & 83.0 & 82.45 & \underline{83.30} & \textbf{83.57} \\
\bottomrule
\end{tabular*}
\end{table}

%% file: reviewer_eleven_dataset_table.tex
\begin{table}[tbp]
\centering
\caption{Benchmark accuracy on eleven datasets, including ImageNet. Accuracy is averaged equally across datasets for each support-sampling seed; entries report the mean $\pm$ sample SD across seeds 1--3. The final row also averages the five shot counts. Gains are paired differences from our paired reproduction of the ProKeR global predictor, in percentage points.}
\label{tab:reviewer-eleven-datasets}
\setlength{\tabcolsep}{2.5pt}
\footnotesize
\begin{tabular*}{\linewidth}{@{\extracolsep{\fill}}rccccc@{}}
\toprule
$K$ & Global control & LAMP & LAMP-F & $\Delta$ LAMP & $\Delta$ LAMP-F \\
\midrule
\multicolumn{6}{c}{ViT-B/16} \\
1 & $71.17 \pm 0.47$ & $72.18 \pm 0.46$ & $73.28 \pm 0.40$ & $1.00 \pm 0.06$ & $2.11 \pm 0.20$ \\
2 & $73.32 \pm 0.49$ & $74.27 \pm 0.43$ & $75.62 \pm 0.45$ & $0.96 \pm 0.10$ & $2.30 \pm 0.10$ \\
4 & $76.26 \pm 0.12$ & $77.70 \pm 0.23$ & $78.80 \pm 0.20$ & $1.45 \pm 0.26$ & $2.54 \pm 0.09$ \\
8 & $79.34 \pm 0.09$ & $80.65 \pm 0.11$ & $80.94 \pm 0.16$ & $1.30 \pm 0.09$ & $1.59 \pm 0.19$ \\
16 & $82.30 \pm 0.09$ & $83.30 \pm 0.15$ & $83.57 \pm 0.13$ & $0.99 \pm 0.08$ & $1.27 \pm 0.08$ \\
Mean & $76.48 \pm 0.09$ & $77.62 \pm 0.12$ & $78.44 \pm 0.12$ & $1.14 \pm 0.03$ & $1.96 \pm 0.03$ \\
\midrule
\multicolumn{6}{c}{RN50} \\
1 & $63.69 \pm 0.23$ & $64.79 \pm 0.32$ & $65.91 \pm 0.15$ & $1.11 \pm 0.23$ & $2.22 \pm 0.35$ \\
2 & $66.42 \pm 0.13$ & $67.50 \pm 0.47$ & $68.83 \pm 0.61$ & $1.07 \pm 0.35$ & $2.41 \pm 0.48$ \\
4 & $69.62 \pm 0.46$ & $71.19 \pm 0.22$ & $72.65 \pm 0.05$ & $1.57 \pm 0.24$ & $3.03 \pm 0.41$ \\
8 & $72.93 \pm 0.10$ & $75.03 \pm 0.07$ & $75.50 \pm 0.09$ & $2.09 \pm 0.10$ & $2.57 \pm 0.19$ \\
16 & $76.62 \pm 0.05$ & $78.73 \pm 0.20$ & $79.10 \pm 0.18$ & $2.11 \pm 0.14$ & $2.48 \pm 0.12$ \\
Mean & $69.86 \pm 0.02$ & $71.45 \pm 0.12$ & $72.40 \pm 0.13$ & $1.59 \pm 0.14$ & $2.54 \pm 0.14$ \\
\bottomrule
\end{tabular*}
\end{table}

%% file: v18_eleven_dataset_effects_table.tex
\begin{table}[tbp]
\centering
\footnotesize
\setlength{\tabcolsep}{3pt}
\caption{Paired accuracy gains at 4 shots on all eleven benchmark datasets, in percentage points. LAMP minus global measures the gain from training-free multi-layer integration; LAMP-F minus LAMP measures the additional gain from prototype adaptation. Entries report the mean $\pm$ sample SD across three support-sampling seeds. For the mean row, dataset gains are first averaged within each seed.}
\label{tab:eleven-dataset-effects}
\begin{tabular*}{\linewidth}{@{\extracolsep{\fill}}lrrrr@{}}
\toprule
 & \multicolumn{2}{c}{LAMP minus global} & \multicolumn{2}{c}{LAMP-F minus LAMP} \\
\cmidrule(lr){2-3}\cmidrule(lr){4-5}
Dataset & ViT-B/16 & RN50 & ViT-B/16 & RN50 \\
\midrule
ImageNet & $+0.168 \pm 0.137$ & $+0.127 \pm 0.030$ & $+0.575 \pm 0.066$ & $+0.835 \pm 0.035$ \\
Caltech101 & $+0.419 \pm 0.328$ & $+0.852 \pm 0.226$ & $+0.203 \pm 0.361$ & $+1.041 \pm 0.366$ \\
Pets & $+0.191 \pm 0.330$ & $+0.145 \pm 0.186$ & $+0.372 \pm 0.576$ & $+0.073 \pm 0.126$ \\
Cars & $+0.858 \pm 0.402$ & $+1.729 \pm 1.296$ & $+1.206 \pm 0.914$ & $+1.190 \pm 0.549$ \\
Flowers102 & $+1.164 \pm 0.394$ & $+2.328 \pm 0.131$ & $+0.392 \pm 0.680$ & $+0.758 \pm 0.587$ \\
Food101 & $+0.232 \pm 0.074$ & $+0.032 \pm 0.046$ & $+0.270 \pm 0.010$ & $+0.901 \pm 0.053$ \\
Aircraft & $+0.160 \pm 0.568$ & $+2.380 \pm 1.291$ & $+0.910 \pm 0.272$ & $+1.330 \pm 0.191$ \\
SUN397 & $+1.547 \pm 2.893$ & $-0.032 \pm 0.023$ & $+5.620 \pm 4.598$ & $+6.715 \pm 0.376$ \\
DTD & $+3.861 \pm 0.034$ & $+1.418 \pm 0.369$ & $+0.355 \pm 0.682$ & $+0.985 \pm 1.731$ \\
EuroSAT & $+5.860 \pm 0.766$ & $+6.695 \pm 1.796$ & $+1.469 \pm 0.460$ & $+0.996 \pm 0.476$ \\
UCF101 & $+1.445 \pm 0.291$ & $+1.568 \pm 0.916$ & $+0.696 \pm 0.313$ & $+1.269 \pm 0.828$ \\
\midrule
Mean & $+1.446 \pm 0.255$ & $+1.568 \pm 0.237$ & $+1.097 \pm 0.289$ & $+1.463 \pm 0.176$ \\
\bottomrule
\end{tabular*}
\end{table}

%% file: imagenet_backbone_table.tex
\begin{table}[tbp]
\centering
\renewcommand{\baselinestretch}{1}\fontsize{9.5}{11.5}\selectfont
\setlength{\tabcolsep}{0pt}
\renewcommand{\arraystretch}{1.10}
\caption{Training-free ImageNet 16-shot accuracy (\%) across four CLIP backbones. External scores come from ProKeR Table~6~\cite{proker} and IDEA Table~2~\cite{idea}. LAMP reports means over seeds 1--3. Bold marks the highest accuracy among the listed methods in each column; gains over ProKeR are in percentage points.}
\label{tab:imagenet-published}
\begin{tabular*}{\linewidth}{@{\extracolsep{\fill}}lrrrr@{}}
\toprule
Models & ResNet-50 & ResNet-101 & ViT-B/32 & ViT-B/16 \\
\midrule
Zero-shot CLIP & 60.33 & 62.53 & 63.80 & 68.73 \\
Tip-Adapter & 61.43 & 64.08 & 65.18 & 70.25 \\
APE & 62.60 & 65.61 & 66.31 & 71.37 \\
IDEA & 62.58 & 65.51 & 65.93 & 71.07 \\
ProKeR & 64.45 & 67.39 & 68.12 & 73.25 \\
\textbf{LAMP (Ours)} & \textbf{65.73} & \textbf{68.52} & \textbf{69.12} & \textbf{74.10} \\
\midrule
Gain over ProKeR (pp) & +1.28 & +1.13 & +1.00 & +0.85 \\
\bottomrule
\end{tabular*}
\end{table}

%% file: rn101_main_table.tex
\begin{table}[tbp]
\centering
\renewcommand{\baselinestretch}{1}\fontsize{9}{11}\selectfont
\setlength{\tabcolsep}{3pt}
\renewcommand{\arraystretch}{1.08}
\caption{Training-free RN101 accuracy averaged equally over the ten non-ImageNet datasets.
LAMP and the global-feature-only baseline use the same support banks. Accuracy is in percent;
gains and pointwise 95\% paired-bootstrap intervals are in percentage points.}
\label{tab:rn101-transfer}
\begin{tabular*}{\linewidth}{@{\extracolsep{\fill}}rrrrl@{}}\toprule
Shot & Global-only & LAMP & Gain & 95\% interval \\\midrule
1 & 65.750 & 66.946 & +1.196 & [+0.873, +1.523] \\
2 & 68.491 & 70.030 & +1.539 & [+1.240, +1.832] \\
4 & 72.333 & 74.146 & +1.814 & [+1.536, +2.104] \\
8 & 75.873 & 78.206 & +2.333 & [+1.989, +2.685] \\
16 & 79.790 & 81.927 & +2.137 & [+1.893, +2.384] \\
\bottomrule\end{tabular*}\end{table}

%% file: selected_weights_figure.tex
\begin{figure}[!tbp]
\centering\includegraphics[width=0.87\linewidth]{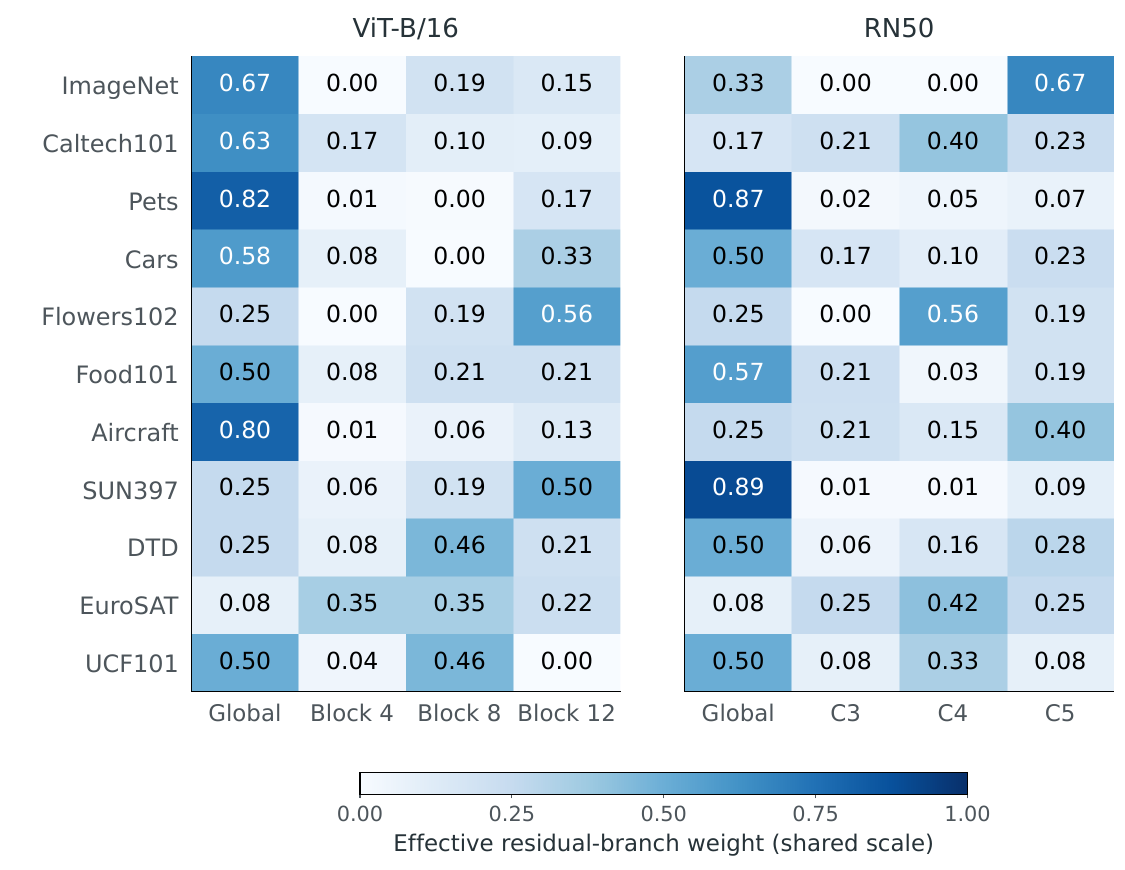}
\caption{Effective weights of the global and regional corrections at 4 shots, averaged over seeds 1--3. The four coefficients in Eq.~\eqref{eq:prism} are $(1-\rho,\rho w_1,\rho w_2,\rho w_3)$; their weighted corrections are added to the shared CLIP prior. Both panels use the same 0--1 scale.}
\label{fig:effective-weights}
\end{figure}

%% file: deployment_cost_main_table.tex
\begin{table}[!t]
\centering
\renewcommand{\baselinestretch}{1}\fontsize{9}{11}\selectfont
\setlength{\tabcolsep}{3pt}\renewcommand{\arraystretch}{1.06}
\caption{Deployment cost at 16 shots, measured on one RTX 4090 with seed 1.
Latency is the median time from image decoding to classification (ms) at
batch size one. Throughput (images/s) and peak allocated GPU memory (GiB)
are measured at batch size 128, separately for each method. Timing includes
decoding, preprocessing, encoding, and classification; disk retrieval is
excluded.}
\label{tab:deployment-cost}
\begin{tabular*}{\linewidth}{@{\extracolsep{\fill}}lllrrr@{}}
\toprule
Dataset & Backbone & Method & Latency & Throughput & GPU peak \\
\midrule
DTD & ViT-B/16 & Global-only & 12.52 & 128.71 & 1.82 \\
 & & LAMP & 42.38 & 124.15 & 2.23 \\
 & & LAMP-F & 42.70 & 127.02 & 2.29 \\
 & RN50 & Global-only & 12.67 & 138.47 & 1.84 \\
 & & LAMP & 23.40 & 137.79 & 2.02 \\
 & & LAMP-F & 23.64 & 137.43 & 2.11 \\
\midrule
ImageNet & ViT-B/16 & Global-only & 14.22 & 113.38 & 2.71 \\
 & & LAMP & 64.83 & 107.89 & 7.64 \\
 & & LAMP-F & 65.68 & 108.05 & 8.56 \\
 & RN50 & Global-only & 14.81 & 118.64 & 3.02 \\
 & & LAMP & 65.86 & 112.28 & 11.55 \\
 & & LAMP-F & 68.04 & 109.82 & 13.39 \\
\bottomrule
\end{tabular*}
\end{table}

%% file: discussion.tex
\section{Discussion}
\label{sec:discussion}

\subsection{Implications for training-free adaptation}
LAMP improves adaptation by combining complementary features from different
encoder layers. The final global descriptor provides image--text alignment,
while intermediate feature maps supply additional visual information for the
target task. Controls with matched search budgets show that descriptors from
multiple layers outperform an ensemble built from the final global feature.
Likewise, pooling several feature maps outperforms pooling only the deepest
map. In the pooling study, regional centroids and whole-map descriptors retain
most of the accuracy gain. These compact representations offer useful options
for applications with limited storage or strict latency requirements.

Calibration and task-specific mixing determine how each layer contributes
to prediction. Calibration is particularly effective for RN50 on EuroSAT,
and the selected layer weights vary across datasets and backbones. LAMP-F
adds prototype adaptation, with the largest gains when few support images
are available. Caching the response to prototype changes makes their evaluation
inexpensive relative to constructing the regional predictors. The results
highlight the value of the visual information already present in a frozen
encoder and show how to use it through complementary residual corrections.

\subsection{Limitations}
\label{sec:limitations}
Using features from multiple layers increases storage requirements and
kernel computation costs. On ImageNet at 16 shots, LAMP increases latency
at batch size one from 14.22 to
64.83\,ms on ViT-B/16 and from 14.81 to 65.86\,ms on RN50
(Table~\ref{tab:deployment-cost}). Dense fitting requires $O(N^2)$ storage
and $O(N^3)$ factorization per branch; LAMP-F also stores prototype responses.

%% file: conclusion.tex
\section{Conclusion}
We presented LAMP, a training-free method that combines global and regional predictions from multiple layers of a frozen vision--language encoder. Each branch retains its native feature dimension and independently fits the same support-label residual; their corrections are then combined in class-score space. LAMP achieves state-of-the-art mean accuracy among the compared training-free RN50 methods at 4, 8, and 16 shots, while matching or exceeding representative methods that require training at higher support budgets. LAMP-F adds text-anchored prototype adaptation and uses a cached response to update the predictions exactly. Beyond these accuracy gains, LAMP provides a way to combine features of different dimensions without learning a shared projection. Its separation of branch fitting, score fusion, and prototype updates offers a foundation for studying task-specific layer selection and developing more scalable residual prediction methods.

%% file: backmatter.tex
\section*{Acknowledgements}
This work was supported by the Young Scientific and Technological Talent Support Program under the Taizhou Fengcheng Talent Plan. The funder provided financial support only and had no role in the study design, conduct of the experiments, data analysis, manuscript preparation, or decision to submit the article for publication.

\section*{Declaration of generative AI and AI-assisted technologies}
During manuscript preparation, the authors used OpenAI ChatGPT/Codex to assist with language editing, document organization, experimental-code preparation, and consistency checks against the recorded experiments. The authors reviewed the material and take responsibility for the article.